# STEEPEST ASCENT HILL CLIMBING
# FOR A MATHEMATICAL PROBLEM

SIBY ABRAHAM[†],	IMRE KISS,	SUGATA SANYAL,	MUKUND SANGLIKAR
Dept of Maths & Stats,	Dept of Eng & Managmt,	School Tech & Comp Sc,	Dept of Mathematics,
G. N. Khalsa College,	Faculty of Engineering	Tata Inst. of Fundamental	Mithibai College,
University of Mumbai,	University Polytechnica	Research, Mumbai,	University of Mumbai,
Mumbai, India.	Timisoara, Romania.	Mumbai, India.	Mumbai, India.
sibyam@gmail.com	imre.kiss@fih.upt.ro	sanyal@tifr.res.in	masanglikar@rediffmail.com

**ABSTRACT**

The paper proposes artificial intelligence technique called hill climbing to find numerical solutions of Diophantine Equations. Such equations are important as they have many applications in fields like public key cryptography, integer factorization, algebraic curves, projective curves and data dependency in super computers. Importantly, it has been proved that there is no general method to find solutions of such equations. This paper is an attempt to find numerical solutions of Diophantine equations using steepest ascent version of Hill Climbing. The method, which uses tree representation to depict possible solutions of Diophantine equations, adopts a novel methodology to generate successors. The heuristic function used help to make the process of finding solution as a minimization process. The work illustrates the effectiveness of the proposed methodology using a class of Diophantine equations given by

$$a_1 \cdot x_1^{p_1} + a_2 \cdot x_2^{p_2} + \ldots\ldots + a_n \cdot x_n^{p_n} = N \text{ where } a_i \text{ and } N \text{ are integers.}$$

The experimental results validate that the procedure proposed is successful in finding solutions of Diophantine Equations with sufficiently large powers and large number of variables.

## 1.0 INTRODUCTION

A Diophantine Equation [Cohen 2007] [Rossen 1987] [Zuckerman 1980] is a polynomial equation, given by

$$f(a_1, a_2, \ldots, a_n, x_1, x_2, \ldots, x_n) = N \qquad (1)$$

where $a_i$ and N are integers. These equations, which were initially studied in detail by third century BC Alexandrian Mathematician Diophantus [Bashmakova 1997] [Bag 1979], have many different types. The simplest ones are the linear equations given by:

$$ax_1 + bx_2 = c \qquad (2)$$

The equations of the form

$$x_1^2 + x_2^2 = x_3^2 \qquad (3)$$

are important as they give solutions, which are Pythagorean triplets. In 1665, French Mathematician Fermat popularized such equations by famously stating that equations of the form

$$x_1^n + x_2^n = x_3^n \qquad (4)$$

have no solutions for n>2, though the world had to wait till 1994 for an actual proof, which used elliptic curves [Shirali & Yogananda 2003]. An elliptic curve (Stroeker and Tzanakis, 1994; Poonen, 2000) is a particular type of Diophantine equation given by

$$y^2 = x^3 + ax + b, \qquad (5)$$

where and b are rational numbers and the right hand side of the equation (5) are given to have distinct roots. There are many such important equations in the collection of Diophantine equations.

___________________________

[†] **CONTACT AUTHOR**

Diophantine equations are used extensively in many fields. Elliptic curve based public key cryptosystems [Lin CH 1995][ Laih CS 1997] [Koblitz 1984] offer better security provisions comparing with other cryptosystems. The performance of super computers can be enhanced by parallelizing compilers to check the problem of data usage which can be reduced to characterization of a Diophantine equation [Zhiyu 1989]. Computable economics [Velu 2004] uses decision problems like Diophantine equations to propose a change in the market equilibrium conditions instead of the conventional parameters. Integer factorization [Knuth 1997] uses Diophantine equations in the process of breaking down a composite number into smaller non-trivial divisors. Diophantine equations are also used in other areas like algebraic curves [Ponnen 2000], projective curves [Brown & Myres 2002] [Stroeker & Tzanakis 1994] and theoretical computer science [Ibarra 2004][Guarari  1982]. These application areas make Diophantine equations an important domain not just in the realm of Mathematics but in other fields too.

Though Diophantine equations have a great historical background and have been used in many areas, there does not exist a general method to find solutions of such equations [Davis 1992] [Matiyasevich 1993].  Then, finding numerical solutions to such equations is the only next way out. This is a tough task as the computing complexity involved in such a process is quite high. In this regard, applying artificial intelligence techniques, which are known for maneuvering huge search space, is significant. Literature talks about few attempts to find numerical solutions of Diophantine equations using hard computing and soft computing techniques of Artificial Intelligence. Abraham and Sanglikar [Abraham and Sanglikar 2001] used basic genetic operators like mutation, inversion and crossover [Michalewich 1992] to find numeric solutions of some elementary equations. They [Abraham and Sanglikar 2007 a] later used a procedure called 'host parasite co-evolution' [Hills 1992] [Paredis 1996][Wiegand 2003] in a typical genetic algorithm to find numerical solutions. They also proposed [Abraham and Sanglikar 2007 b] a unique evolutionary and co-evolutionary [Rosin and Belew 1997] computing method to find numerical solutions of such equations. Joya et al [Joya et al., 1991] used higher order Hopfield neural networks to find solutions of Diophantine equations.  Abraham and Sanglikar [Abraham and Sanglikar 2008] offered simulated annealing as a possible strategy to find solutions of these equations. Abraham *et al* [Abraham *et al* 2010] discussed in detail a particle swarm optimization based method to find numerical solutions of such equations. In addition to these methods based on soft computing, literature also mentions A * search based hard computing mechanism as a possible alternative to find numerical solutions of these equations [Abraham and Sanglikar 2009]. Hard computing methods are significant as they try to explore as many candidate solutions as possible in a systematic way unlike soft computing, which uses randomness in the process and hence risks of 'slipping away' the solutions on the way.

This paper proposes hill climbing as a hard computing artificial intelligence technique to find numerical solutions of Diophantine equations. Hill Climbing is a local search [Russel & Norwig 2003] technique. It starts with an initial solution and steadily and gradually generates neighboring successor solutions. If the neighboring state is better than the current state, we make the neighboring state the current state. The whole process can be taken as an optimization process [Lugar 2006]. There are different variants of hill climbing. They are simple hill climbing, steepest hill climbing, stochastic hill climbing and random restart hill climbing. The paper uses steepest ascent version of the hill climbing to find numerical solution of Diophantine equations. In steepest hill climbing all successor nodes are probed and compared for its relevance and then the best amongst them is taken as the successor node. This results in having an exhaustive local search and identification of the best possible successor of a given node at any instant of time.

**2.0 HILL-DOES Methodology:** The system developed to find numerical solutions of Diophantine equations using Hill climbing is based on the Steepest Ascent version of Hill Climbing and is called HILL-DOES. It uses a system of equations given by

$$a_1 \cdot x_1^{p_1} + a_2 \cdot x_2^{p_2} + \ldots\ldots + a_n \cdot x_n^{p_n} = N \qquad (6)$$

where $a_i$ and $N$ are integers, for demonstrating the effectiveness of the system proposed.

**2.1 Representation:** The possible solutions of the Diophantine equation (6) are represented by a tree whose nodes are taken as n-vectors given by $(x_1, x_2, ..., x_n)$. The procedure starts with an initial solution, given to be (1, 1, 1, ....., 1) and uses two queues in its construction. The first queue, which is called PROBE-Q, is used to store the nodes, which have been probed. The second queue, which is referred as NOPROBE-Q, is used to store the nodes, which have been generated but not better than the current node. These two queues help to separate the generated nodes into two distinct classes – 'probed nodes' and 'not-probed nodes'.

**2.2 Successor nodes:** Successor nodes of the current node are generated in HILL-DOES using specially defined production rules. The production rules applied are given by:

$$(x_1, x_2, ..., x_i, ...., x_n) \rightarrow (x_1, x_2, ..., x_{i+1}, ...., x_n) \text{ for } i = 1, 2, ...., n \qquad (7)$$

These production rules help to generate all possible nodes in the vicinity of the current node. Hill climbing, being a local search technique, needs to explore all possible nodes within the neighborhood of the current node. The successor nodes generated in this way, take care of this requirement of the search strategy.

**2.3 Heuristic function:** The heuristic function used to evaluate the effectiveness of a node $(x_1, x_2, x_3, ...., x_n)$ in the search process is given by

$$H(x_1, x_2, ..., x_n) = N - (a_1 x_1^{p_1} + a_2 x_2^{p_2} + a_3 x_3^{p_3} + \ldots\ldots + a_n x_n^{p_n}) \qquad (8)$$

Since the objective of the procedure is to find numerical solutions of equation (6), the problem reduces to find a vector given by $(x_1, x_2, ..., x_n)$ with $H(x_1, x_2, ..., x_n) = 0$. The value of 'H' shows how far is a given node away from the goal node. Lower the value of 'H', closer is the node to the solution. However, the negative value of 'H' requires some extra care to make the search process on track. Whenever H value becomes negative, the proposed procedure does not expand the corresponding node even if that has better heuristic function value compared to others. Instead, the node with the next better heuristic function value is expanded and the process is continued. In other words, the nodes having negative H values are replaced with the better nodes from the NO-PROBEQ.

**2.4 Backtracking**: It is possible to have a current node, with all its successor nodes having inferior heuristic values in comparison with that of the current node. Steepest ascent hill climbing always demands having better nodes as successor nodes to continue the procedure. This drawback of hill climbing is overcome in the procedure by incorporating a strategy of backtracking. As per this, when the procedure fails to produce better nodes as successor nodes, it leaves the current node and goes back or backtracks to the previous best node generated. Then, the exploration process resumes from that node and the process of traversing through the tree in another path is followed. It is quite possible that during the search process, the procedure might hit on such inferior successor nodes on a regular basis. At these instances, the procedure is continued with backtracking at each and every instance. This way an unhindered search procedure is guaranteed always in HILL-DOES.

**2.5 Algorithm:** The basic steepest ascent hill climbing algorithm is slightly restructured to be acquainted with the constraints of Diophantine equations. The algorithm used in HILL-DOES is explained in the following lines.

    **Step1**: Initialize node, which is usually (1, 1, . . . . . , 1). Evaluate it.
        Put it in the PROBE-Q.
    **Step2**: If PROBE-Q is empty, then stop.
    **Step3**: Pick first node from PROBE-Q. Label it as current node: C-Node.
    **Step4**: If C-Node is the goal node then return C-Node as a solution.
        (Goal state is reached when $H(x_1, x_2, ....., x_n) = 0$
    **Step5**: If C-Node is not a goal node, check whether $H(x_1, x_2, ....., x_n) > 0$
        a) If yes, generate successors of C-Node and evaluate them.
        b) If no, Go to step 2.
    **Step6**: Compare successors of C-node and the better node amongst them.
        Store the remaining nodes in the NOPROBE-Q.
        a) If the better node is better than C-Node and if it has not been probed before, then make it C-Node. If it has been probed, then pick the first element from NOPROBE-Q and make it as the C-Node. Put C-Node in the PROBE-Q.
        b) If the better node is not better than C-Node, then pick the first element from the NOPROBE-Q, make it as the C-Node and put in PROBE-Q. If NOPROBE-Q is empty, then do nothing.
    **Step6**: Go to step 2.

The algorithm as it is used in HILL-DOES is illustrated using a simple Diophantine Equation given by $x_1^2 + x_2^2 = 100$. The initial node is (1, 1) and $H(1, 1) = 100 - (1^2 + 1^2) = 98$.
    **Step1**: Put (1, 1) in PROBE-Q.
    **Step3:** C-Node = (1,1).
    **Step5**: Since it is not a goal node and H(C-Node) = 98 > 0, generate Successors of it.

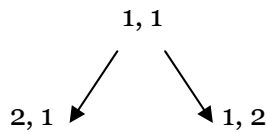

    **Step6**: Both (2,1) and (1, 2) have the same heuristic function value 95. So, choose any one say (2,1). Put (2,1 ) in PROBE-Q and (1, 2) in NOPROBE-Q. Generate children of (2,1).

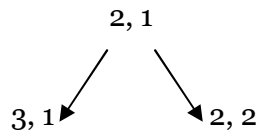

Since the heuristic value of node (3, 1) is 90, which is better than the heuristic value 92 of the node (2, 2) and 95 of (2, 1) we select (3, 1) as the next node to be expanded. Continue this process of generating successors and identifying the best amongst them to be C- Node, which is illustrated in figure 1. If the process gets stuck in not finding a better successor, back track to the previous node and continue the process of exploration until a node with heuristic function value zero is generated. Such a node will be the solution of the given Diophantine equation.

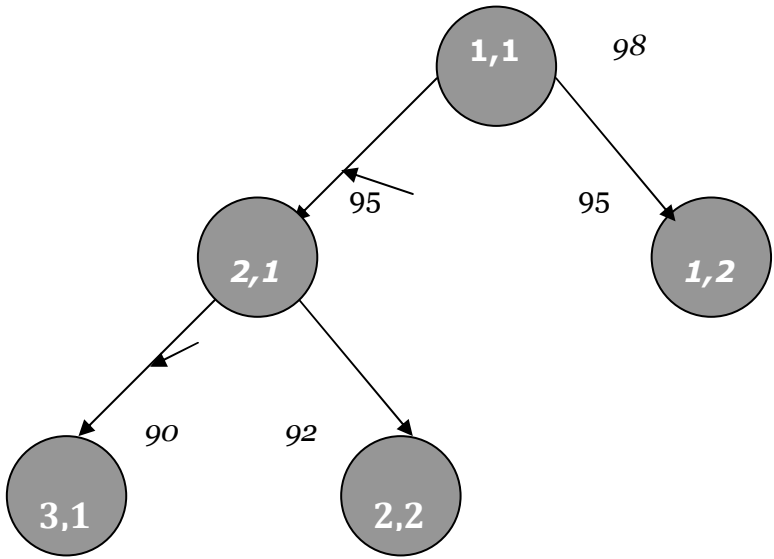

Figure 1: Search tree of $x_1^2 + x_2^2 = 100$

**3.0 ANALYSIS, DISCUSSION AND INTERPRETATION:** The procedure discussed in HILL-DOES has been implemented in Java. The user supplies the details of the equation like number of variables involved, coefficients, powers and the value of N. The experimental results have been analyzed and discussed in the following sections.

**3.1 Nodes Generated:** Figure 2 shows the nodes generated by HILL-DOES for an elementary equation $x_1^2 + x_2^2 = 149$, before finding the first solution (10,4). The process generated 68 nodes during the search process. The figure shows the steady search of the process in the search space. Figure 3 demonstrates the convergence of heuristic function values of the nodes generated in the same demonstration. Initially, there is a sudden reduction of heuristic function values and once the process becomes mature, there is a directed approach towards the value zero, finally resulting in the solution.

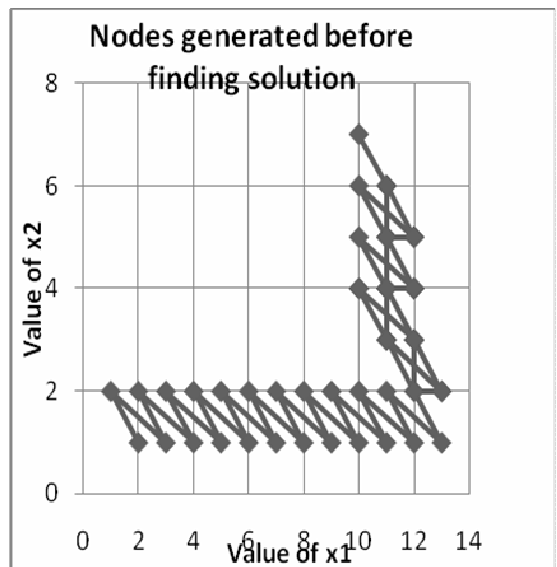

Figure 2: Nodes generated during the search

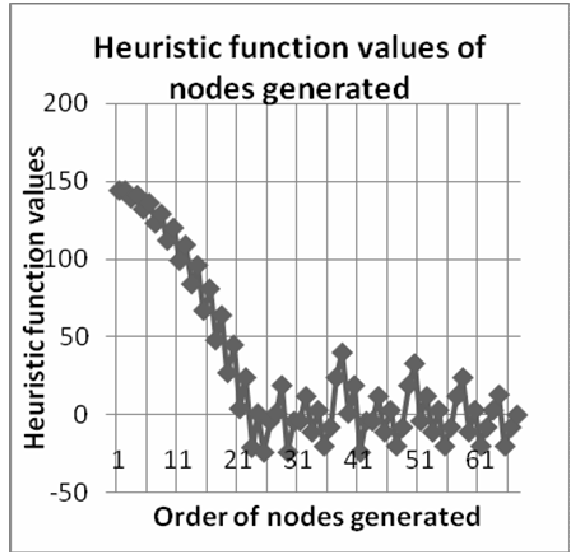
Figure 3: Heuristic function values of nodes

**3.2 Results on equations with varying degrees:** Table 1 demonstrates the results obtained when the system was run for different Diophantine equations with varying values for the degrees. It shows that irrespective of reasonably large values for degrees and higher values of N, the system could give solutions within a smaller number of iterations. This points out that those large values of N do not affect the efficiency of the system. In addition, the comparatively lesser number of iterations only consumed for finding the solutions also validate the effectiveness of the system in finding solutions of Diophantine equations with larger value of N.

| Sr. No | Diophantine Equation | Degree of equation | Solution Found | Iterations |
|---|---|---|---|---|
| 1 | $x_1^2 + x_2^2 = 625$ | 2 | 24, 7 | 29 |
| 2 | $x_1^3 + x_2^3 = 1008$ | 3 | 10, 2 | 10 |
| 3 | $x_1^4 + x_2^4 = 1921$ | 4 | 6, 5 | 9 |
| 4 | $x_1^5 + x_2^5 = 19932$ | 5 | 7, 5 | 10 |
| 5 | $x_1^6 + x_2^6 = 47385$ | 6 | 6, 3 | 7 |
| 6 | $x_1^7 + x_2^7 = 4799353$ | 7 | 9, 4 | 11 |
| 7 | $x_1^8 + x_2^8 = 16777472$ | 8 | 8, 2 | 8 |
| 8 | $x_1^9 + x_2^9 = 1000019683$ | 9 | 10, 3 | 11 |
| 9 | $x_1^{10} + x_2^{10} = 1356217073$ | 10 | 8, 7 | 13 |

Table 1: Results on equations with varying degrees

**3.3 Results on equations with varying number of variables:** Table 2 shows the results obtained when HILL-DOES was run on Diophantine equations with varying number of variables. This shows that the system provides solutions even when the number of variables is competitively high.

| Sr. No | Diophantine Equation | No. of variables | Solution Found | Iteration required. |
|---|---|---|---|---|
| 1 | $x_1^2+x_2^2 = 149$ | 2 | 10, 7 | 34 |
| 2 | $x_1^2+x_2^2+x_3^2 = 230$ | 3 | 15, 2, 1 | 15 |
| 3 | $x_1^2+x_2^2+...+x_4^2= 295$ | 4 | 17, 2, 1, 1 | 17 |
| 4 | $x_1^2+x_2^2+....+x_5^2= 325$ | 5 | 17, 1, 1, 3, 5 | 22 |
| 5 | $x_1^2+x_2^2+....+x_6^2= 420$ | 6 | 20, 1, 1, 1, 1, 4 | 22 |
| 6 | $x_1^2+x_2^2+....+x_7^2= 450$ | 7 | 21, 2, 1, 1, 1 ,1 | 21 |
| 7 | $x_1^2+x_2^2+....+x_8^2= 590$ | 8 | 23, 2, 1, 1, 1,1, 2, 7 | 86 |
| 8 | $x_1^2+x_2^2+....+x_9^2= 720$ | 9 | 26, 2, 1, 1, 1, 2, 2, 2, 5 | 42 |
| 9 | $x_1^2+x_2^2+....+x_{10}^2=956$ | 10 | 30, 2, 1, 1, 1,1, 2, 2, 2, 6 | 48 |

Table 2: Results on equations with varying number of variables

**3.4 Conclusion:** The paper presents steepest ascent hill climbing search based procedure to find numerical solution of Diophantine equations. Local optimum points were tackled by resorting to backtracking as and when the procedure hit on such local optimum points. The experimental results showed that the technique work fine for Diophantine equations of varied types. However, the solutions generated, especially when the number of variables is large, have the tendency to have the coordinates closely placed. Further enhancement to the work is directed at addressing this issue.

**5. REFERENCE:**
[Abraham & Sanglikar 2001]: Abraham, S and Sanglikar, M; 'Diophantine equation solver-a genetic algorithm application', Mathematical Colloquium Journal, Vol. 15, No 3, pp 16-20, 2001.
[Abraham & Sanglikar 2007a]: Abraham, S and Sanglikar, M; 'Nature's way of avoiding premature convergence: a case study of Diophantine equations', Proceedings of the International Conference on Advances in Mathematics: Historical Developments and Engineering Applications, Pantnagar, Uttarakhand, India, 19–22 December, pp 182, 2007.
[Abraham & Sanglikar 2007b]: Abraham, S and Sanglikar, M; 'Finding solution to a hard problem: an evolutionary and co-evolutionary approach' Proceedings of the International Conference on Soft Computing and Intelligent Systems, Jabalpur, India,27–29 December, pp 262-267, 2007.
[Abraham & Sanglikar 2008]: Abraham, S and Sanglikar, M; 'Finding numerical solution to a Diophantine equation: simulated annealing as a viable search strategy', Proceedings of the International Conference on Mathematical Sciences, United Arab Emirates University, Al Ain, UAE, 3–6, pp 319, March, 2008.
[Abraham & Sanglikar 2009]: Abraham, S and Sanglikar, M; 'A* search for a challenging problem', Proceedings of the 3rd Internl Confer.on Maths and Computer Science, Loyola College, Chennai, 5th–6th January, pp 453-457, 2009.
[Abraham et al 2010]: Abraham, S; Sanyal, S and Sanglikar, M; 'Particle Swarm Optimization based Diophantine Equation Solver', International Journal of Bio-inspired Computation, Vol 2, No 2, pp 100-114, 2010.

[Bag 1979]: Bag A K; 'Mathematics in ancient and medieval India', Chaukhambha Orientalia, Delhi, 1979.

[Bashmakova 1997]: Bashmakova et al; 'Diophantus and Diophantine Equations, Mathematical Association of America, 1997.

[Brown & Myers 2002]: Brown, E. and Myers, B; 'Elliptic curves from Model to Diophantus and Back', The Mathematical Association of America Monthly, August–September, Vol. 109, pp.639–649, 2002.

[Cohen 2007]: Cohen, H; Number Theory, Vol. I: Tools and Diophantine Equations and Vol. II: Analytic and Modern Tools, Springer-Verlag, pp. 239- 240, 2007.

[Davis 1982]: Davis, M; 'Hilbert's tenth problem is unsolvable', Computability and Unsolvability, Appendix 2, 1999-235, Dover, New York, 1982.

[Guarari 1982]: Guarari, E.M; 'Two way counter machines and Diophantine equations', Journal of ACM, Vol. 29, No. 3, 1982.

[Hills 1992]: Hills, W.D; 'Co-evolving parasites improve simulated evolution as an optimization procedure', Artificial Life, Addison Wesley, Vol. 2, 1992.

[Ibarra et al. 2004]: Ibarra, O.H. et al; 'On two way FA with monotone counters and quadratic Diophantine equations', Theoretical Computer Science, Vol. 312, pp.2–3, 2004.

[Joya et al 1991]: Joya et al; 'Application of Higher order Hopfield neural networks to the solution of Diophantine equation', Lect. Notes in Comp Sc, Vol.540, Springer, 1991.

[Knuth D 1997]: Knuth, D; 'The Art of Computer Programming, Volume 2: Semi-numerical Algorithms', Third Edition. Addison-Wesley, 1997.

[Koblitz 1984]: Koblitz, N; Introduction to Elliptic Curves and Modular Forms, Springer, 1984.

[Laih. et al., 1997]: Laih, C.S. et al; 'Cryptanalysis of Diophantine equation oriented public key cryptosystem', IEEE Transactions on Computers, April, Vol. 46, 1997.

[Lin et al. 1995]: Lin, C.H. et al 'A new public-key cipher system based upon Diophantine equations', IEEE Transactions on Computers, January, Vol. 44, 1995.

[Luger 2006]: Luger, G.L; Artificial Intelligence: Structures and Strategies for Complex Problem Solving, 4e Pearson Education, 2006.

[Matiyasevich 1993]: Matiyasevich, Y.V; Hilbert's Tenth Problem, MIT press, 1993.

[Michalewich 1992]: Michalewich, Z; GA + Data Structures = Evaluation Programs, Springer Verlag, 1992.

[Paredis, J. 1996]: Paredis, J; 'Co-evolutionary computation', Artificial Life Journal, Vol. 2, No. 3, 1996.

[Poonen 2000]: Poonen, B; 'Computing rational points on curves', Proceedings of the Millennial Conference on Number Theory, 21–26 May, University of Illinois at Urbana-Champaign, 2000.

[Rosin & Belew 1997]: Rosin, C.D. and Belew, R.K; 'New methods for competitive co-evolution', Evolutionary Computation, Vol. 5, No. 1, pp.1–29, 1997.

[Rossen, K. 1987]: Rossen, K; Elementary Number Theory and its Applications, Addison, 1987.

[Russell &Norwig 2003]: Russell, S. and Norwig, P; Artificial Intelligence– A Modern Approach, 2nd ed., Pearson, 2003.

[Shirali & Yogananda 2003]: Shirali, S. and Yogananda, C.S; 'Fermat's last theorem: a theorem at last', Number Theory: Echoes From Resonance, University Press, 2003.

[Stroeker & Tzanakis 1994]: Stroeker & Tzanakis,"Solving elliptic Diophantine equations by estimating linear forms in elliptic logarithms, Acta Arithmetica, Vol. 67, No. 2, 1994.

[Velu 2000]: Velu Pillai K; Computable Economics: The Arne Memorial Lecture Series, Oxford University Press, 2000.

[Wiegand 2003]: Wiegand, P; 'An analysis of cooperative co-evolutionary algorithms', PhD dissertation, George Mason University, 2003.

[Zhiyu 1989]: Zhiyu, S., Li, Z., Yew, P.C. 'An empirical study on array subscripts and data dependencies', CSRD Report 840, August, 1989.

[Zuckerman 1980]: Zuckerman; An Introduction to the theory of numbers', 3rd ed., Wiley, 1980.